\documentclass[twoside,11pt]{article}
\usepackage{blindtext}
\usepackage[T1]{fontenc}
\usepackage{jmlr2e}
\usepackage[utf8]{inputenc}
\usepackage{enumitem}
\usepackage{subcaption} 

\usepackage{makecell}

\usepackage{amsmath} 
\usepackage{amsfonts} 
\usepackage{mathtools} 
\usepackage{bm}
\usepackage[thinc]{esdiff}

\usepackage{algpseudocode} 
\usepackage{xcolor}

\definecolor{mygreen}{RGB}{0,128,0}
\definecolor{cobalt}{rgb}{0.0, 0.28, 0.67}

\usepackage{hyperref} 
\usepackage{tikz}

\usepackage{graphicx}%
\usepackage{multirow}%
\usepackage{amsmath,amsfonts}%
\usepackage{mathrsfs}%
\usepackage[title]{appendix}%
\usepackage{xcolor}%
\usepackage{textcomp}%
\usepackage{manyfoot}%
\usepackage{booktabs}%
\usepackage[ruled,vlined]{algorithm2e}
\SetKwInput{KwInput}{Input}       
\SetKwInput{KwInit}{Initialize}   
\usepackage{algpseudocode}

\usepackage{listings}%
\usepackage{enumitem}

\usepackage{tikz}
\usepackage{algpseudocode} 
\usepackage{subcaption} 

\usepackage{makecell}
\usepackage{bm}

\begin{document}

\title{Shrinkage prior for implicit generative models}

\author{\name Shreya Sinha Roy \email shreya.sinha-roy@warwick.ac.uk\\
\name Christian P. Robert\thanks{Also affiliated with CEREMADE, Université Paris Dauphine PSL, France} \email c.a.m.robert@warwick.ac.uk\\
\name Ritabrata Dutta \email ritabrata.dutta@warwick.ac.uk\\
\addr Department of Statistics, University of Warwick\\
Coventry, CV4 7AL\\
United Kingdom}

\editor{} 

\maketitle

\begin{abstract}


Implicit generative models based on deep neural networks have revolutionized many scientific fields, from weather research to language generation. They successfully model complex random objects arising in real-world problems. In this work, we propose a sample-efficient training mechanism for these models. Our approach leverages regularized inference with shrinkage priors and generalized Bayes. In the large $p$ (parameter dimension) and small $n$ (sample size) setting, sparsity-inducing priors have been successfully used for linear regression when $p \gg n$. However, their application to implicit generative models remains challenging due to the intractability
of the likelihood function. To address this, we adapt a scoring rule posterior based on a strictly proper scoring rule, following the generalized Bayes framework. We then propose a block SGMCMC within Gibbs sampling mechanism. This enables efficient exploration of high-dimensional parameter spaces by identifying a sparse representation of neural implicit generative models. We demonstrate the performance of our proposed method through experiments on $p \gg n$ linear regression and an application of neural generative models.


\end{abstract}

\section{Introduction}

Implicit generative models are increasingly used to model complex random objects arising in scientific and real-world applications, including weather forecasting \citep{price2024gencastdiffusionbasedensembleforecasting}, population genetics \citep{yelmen2023overview, huang2024harnessing}, protein structure and molecular design \citep{anand2022protein}, and image generation \citep{goodfellow2014generative,ho2020denoising}. Their ability to generate realistic samples without requiring an explicit likelihood has made them a powerful modelling paradigm. However, the lack of a tractable likelihood also creates significant challenges for statistical inference and model assessment.

In this work, we consider statistical inference for such models. Implicit generative models introduce a latent random variable $\mathbf{z} \in \mathbb{R}^k \sim q(\mathbf{z})$, where $q$ is a fixed, tractable distribution on $\mathbb{R}^k$, together with a deterministic nonlinear map $G_{\mathbf{\theta}}: \mathbb{R}^k \rightarrow \mathbb{R}^d$, parameterized by $\mathbf{\theta} \in \mathbb{R}^p$. The model generates samples through
\[
\mathbf{x}=G_{\mathbf{\theta}}(\mathbf{z}),
\]
with distribution
\[
P_{\mathbf{\theta}} = G_{\mathbf{\theta}}\sharp q,
\]
the pushforward distribution of $q$. While traditional approaches often focus on scientific models that incorporate prior knowledge of the underlying process, we consider high-dimensional implicit models where the data-generating mechanism is unknown. In particular, we focus on deep generative models, where $G_{\mathbf{\theta}}$ is represented by a neural network \citep{Mohamed_Lakshminarayanan_2017}. For conditional tasks, where the objective is to model random objects given an observed variable $\mathbf{w} \in \mathbb{R}^m$, we instead consider the map $G_{\mathbf{\theta}}:\mathbb{R}^k \times \mathbb{R}^m \rightarrow \mathbb{R}^d$, which generates
\[
\mathbf{x}=G_{\mathbf{\theta}}(\mathbf{z},\mathbf{w}),
\]
with conditional distribution $P_{\mathbf{\theta}}(\cdot \mid \mathbf{w})$, as in conditional GANs \citep{mirza2014conditional}.

When $G_{\mathbf{\theta}}$ is represented by a large neural network and trained on an observed dataset $\mathcal{D}_n=\{y_1,\ldots,y_n\}$ with limited sample size, statistical inference becomes a large-$p$, small-$n$ problem ($p \gg n$). Bayesian methods based on shrinkage priors are well established for this setting in linear regression \citep{bhadra2019lasso}. However, these methods cannot be applied directly to implicit generative models because the probability density of $P_{\mathbf{\theta}}$ is unavailable. Building on the generalized Bayesian framework of \citet{pacchiardi2024generalized}, which replaces the likelihood with a scoring rule, we develop a sparsity-inducing generalized Bayesian approach using the horseshoe prior \citep{carvalho2010horseshoe}. We further introduce a blocked Gibbs sampler based on preconditioned adjusted stochastic gradient Langevin dynamics (prec-adjSGLD) to efficiently sample from the resulting high-dimensional posterior. To the best of our knowledge, while Bayesian regularization has been extensively studied for deterministic neural networks \citep{gal2016dropout, ghosh2019model, mohammed2023comprehensive}, this is the first work to develop a sparsity-inducing Bayesian framework for neural implicit generative models.

We first study the large-$p$, small-$n$ sparse linear regression problem. Although standard Bayesian methods are available, Markov chain Monte Carlo algorithms are often prohibitively expensive in this setting \citep{johndrow2020scalable}. Our proposed method achieves strong sparsity recovery while providing the best predictive performance at a substantially lower computational cost.
We also consider a sample-efficient training of a conditional implicit generative model for the Hopper, a reinforcement learning benchmark (from the OpenAI Gymnasium package \citep{kwiatkowski2024gymnasiumstandardinterfacereinforcement}), which mimics a Markov decision process.

The rest of the chapter is structured as follows. Section~\ref{ChapHSsec:genbayeshorse} defines a generalized posterior over the parameters of an implicit generative model using the adaptive shrinkage horseshoe prior. We then introduce an approximate blocked Gibbs sampling scheme to sample from the proposed posterior. Section~\ref{ChapHSsec:simulations} presents simulation studies demonstrating the application and advantages of the proposed sampler over fixed priors. We consider two examples: high-dimensional sparse linear regression, and a conditional deep generative model for the reinforcement learning environment Hopper. Finally, Section~\ref{ChapHSsec:conclusion} concludes the chapter.




\section{Generalized Bayesian regularization for implicit generative neural models}\label{ChapHSsec:genbayeshorse}


Due to the intractibility of the likelihood functions for the implicit generative neural models, they are trained by matching the pushforward distribution $P_{\mathbf{\theta}}$ to the data distribution using adversarial objectives \citep{Goodfellow_Pouget-Abadie_Mirza_Xu_Warde-Farley_Ozair_Courville_Bengio_2014}, $f$-divergences \citep{Nowozin_Cseke_Tomioka_2016}, proper scoring rules \citep{pacchiardi2020score, heng2024diffusion} or using prequential scoring rules \citep{pacchiardi2024probabilistic} for conditional generative models. Despite their empirical success, such training procedures are data-intensive and provide limited quantification of uncertainty. This is specially problematic in over-parameterized settings or when data are scarce, well known from many studies for $p\gg n$ sparse linear regression \citep{donoho2009observed}. 

To account for parameter uncertainty in this setup, we propose to adopt a Bayesian approach.
Let $\mathcal{G} = \{G_{\mathbf{\theta}} : \mathbf{\theta} \in \Theta \subseteq \mathbb{R}^p\}$ denote the model class and place a prior distribution $\pi(\mathbf{\theta})$ over parameters. 
Given observed data $\mathcal{D}_n$ and a likelihood function $l(\mathbf{\theta};\mathcal{D}_n)$ under the model $G_{\mathbf{\theta}}$, Bayesian inference yields the posterior distribution
\[
\pi(\mathbf{\theta} \mid \mathcal{D}_n)
=
\frac{\pi(\mathbf{\theta}) l(\mathbf{\theta};\mathcal{D}_n)}
{\int_{\mathbf{\theta}} \pi(\mathbf{\theta}) l(\mathbf{\theta};\mathcal{D}_n) \, d\mathbf{\theta}}.
\]
In the absence of a tractable likelihood \(l(\theta;\mathcal{D}_n)\), we follow \citet{pacchiardi2024probabilistic} and replace the likelihood with the exponential of the negative of a strictly proper scoring rule, yielding a scoring-rule posterior. To further regularize the model, we place a sparsity-inducing horseshoe prior \citep{carvalho2010horseshoe,ghosh2019model} on \(\theta\). We discuss the details in the following sections.


\subsection{Generalized Bayesian likelihood-free Inference}


Likelihood-free Bayesian methods, such as approximate Bayesian computation (ABC) \citep{Beaumont_Zhang_Balding_2002} and neural network–based surrogate approaches \citep{Lueckmann_Goncalves_Bassetto_Öcal_Nonnenmacher_Macke_2017} often consider low-dimensional models with low dimensional parameters. However, they struggle to scale to the high-dimensional parameter spaces typical of deep generative models. Motivated by generalized Bayesian framework with loss functions \citep{bissiri2016general, jewson2018principles, knoblauch2022optimization}, \citep{pacchiardi2024generalized}, we propose to define scoring rule posterior by choosing proper scoring rules \citep{dawid2007geometry} as the loss function. These scoring rules can be unbiasedly estimated from the easily available simulations of the implicit generative models, as shown in the previous chapters.

Scoring rules \citep{gneiting2007strictly} were originally proposed as a diagnostic of a probabilistic forecast, which provides a penalty ($S(P,Y)$) to the forecast ($P$) evaluated at an observed outcome ($Y$). If $Y \sim Q$ denotes the true data-generating distribution, the expected score is defined as $\mathbf S(P,Q) := \mathbb{E}_{Y \sim Q}[S(P,Y)]$. A scoring rule is proper with respect to a class of distributions $\mathcal{P}$ if
\[
\mathbf S(Q,Q) \leq \mathbf S(P,Q), \quad \forall P,Q \in \mathcal{P},
\]
which ensures that truthful forecasting is optimal.
A scoring rule is said to be strictly proper if $\mathbf S(P,Q)$ is uniquely minimized when $P = Q$, the true data-generating distribution.
Strictly proper scoring rules induce statistical divergences between probability measures \citep{dawid2007geometry}.

Given data $\mathcal{D}_n = \{y_1,\dots,y_n\}$, an implicit generative model $G_\mathbf{\theta}$, a prior $\pi(\mathbf{\theta})$, and a weight parameter $\omega>0$, the scoring-rule posterior \citep{pacchiardi2024generalized} becomes
\begin{eqnarray}
\label{ChapHSeq:gen_post}
\pi_S(\mathbf{\theta} \mid \mathcal{D}_n)
\propto
\pi(\mathbf{\theta})\exp\!\left\{-\omega\sum_{i=1}^n S(P_\mathbf{\theta},y_i)\right\},
\end{eqnarray}
where $P_\mathbf{\theta}$ corresponds to the pushforward distribution induced by $G_\mathbf{\theta}$ and, $\omega > 0$ is called the learning rate which determines the influence of the loss with respect to the prior. $\omega$ can be selected to achieve different goals (often, to match some frequentist property, see \citealp{lyddon2019general, syring2019calibrating, matsubara2022generalised}, or to ensure two different scoring-rule posteriors have same scale \citep{pacchiardi2024generalized}). However, we keep $\omega = 1$ fixed for our simulation studies.

When a proper scoring rule is used to define the posterior in Equation~\eqref{ChapHSeq:gen_post}, the resulting distribution satisfies a Bernstein–von Mises–type result under mild regularity conditions \citep{pacchiardi2022statistical}, which we state here. 
\begin{theorem}\label{ChapHSTh:asymptotic_short}
    [\citep{pacchiardi2022statistical}] Suppose, $Q$ is the true underlying data-generation distribution, meaning that $\mathcal{D}_n \sim Q$. Assume the expected scoring rule $\mathbf S(P_\theta, Q) $  has a unique minimizer $ \theta^\star $ and the prior $\pi(\theta)$ is continuous and positive at $ \theta^\star $. Further, denote now by $ \pi_S^{*}\left(\cdot | \mathcal{D}_n\right) $ the density of $ \sqrt{n}\left(\theta-\hat{\theta}^{(n)}\left(\mathcal{D}_n\right)\right) $ when $ \theta\sim\pi_S\left(\cdot | \mathcal{D}_n\right) $, where $  \hat{\theta}^{(n)}\left(\mathcal{D}_n \right) $ is a sequence which converges almost surely to $ \theta^\star $ as $ n\to\infty $. 
    Then, under some technical assumptions, as $n \rightarrow \infty$, with probability 1 over $ \mathcal{D}_n $:
		$$
		\int_{\mathbb{R}^p}\left|\pi_S^{*}\left(s | \mathcal{D}_n\right)-\mathcal{N}\left(s|0,H_\star^{-1}\right)\right| d s  \to 0,
		$$
		where $ \mathcal{N}(\cdot|0, \Sigma) $ denotes the density of a multivariate normal distribution with zero mean vector and covariance matrix $ \Sigma $. Here $H_\star = \nabla_{\theta}^2 \mathbf S(P_{\theta}, Q)|_{\theta = \theta^*}$ is the asymptotic covariance matrix.    
    \end{theorem}
In particular, the scoring-rule posterior concentrates almost surely on the parameter value that minimizes the expected scoring rule, provided the minimizer is unique.
In well-specified models with strictly proper scoring rules, this minimizer coincides with the true parameter value, yielding posterior consistency.
Uniqueness of the minimizer, however, may also hold under model misspecification 
or for non-strictly proper scoring rules. 

To facilitate computation in the likelihood-free scenario, we should choose a strictly proper scoring rule for which we can easily compute an unbiased estimate. From many alternative scoring rules \citep{gneiting2007strictly}, here we choose energy score \citep{rizzo2016energy} defined as
\begin{align*}
S_{\mathrm{E}}^{(\beta)}(P_\mathbf{\theta},y)
&=
2\,\mathbb{E}\!\left[\|X-y\|_2^\beta\right]
-
\mathbb{E}\!\left[\|X-X'\|_2^\beta\right],
\end{align*}
where $X \perp X' \sim P_\mathbf{\theta}$ and $\beta \in (0,2)$. In this work, we will keep $\beta$ fixed at $1$.  
It is strictly proper for distributions satisfying $\mathbb{E}_{X\sim P_\mathbf{\theta}}\|X\|^\beta < \infty$. Given samples $\mathbf{x}_m=\{x_j\}_{j=1}^m$ with $x_j \sim P_\theta$, the energy score can be unbiasedly estimated as,
\begin{align*}
\hat{S}_{\mathrm{E}}^{(\beta)}(\mathbf{x}_m,y)
&=
\frac{2}{m}\sum_{j=1}^m \|x_j-y\|_2^\beta
-\frac{1}{m(m-1)}
\sum_{\substack{j,k=1 \\ j\neq k}}^m
\|x_j-x_k\|_2^\beta.
\end{align*}

\subsection{Sparsity inducing horseshoe prior}

Most works applying Bayesian inference for deterministic neural networks \citep{fortuin2022bayesian}, adopt simple isotropic Gaussian priors or heavy-tailed t-distribution. However, the ``Bad Prior Hypothesis'' \citep{Wenzel_Roth_Veeling_Świątkowski_Tran_Mandt_Snoek_Salimans_Jenatton_Nowozin_2020} highlights the limitations of such priors in overparameterized models, which can be especially hazardous in data-scarce regimes. Gaussian priors implicitly correspond to $\ell_2$-type regularization \citep{bhadra2019lasso}, which may be insufficient to control model complexity in high-dimensional settings.

To address this, shrinkage priors have been proposed as alternatives.
One class consists of discrete mixture priors, such as spike-and-slab \citep{johnstone2004needles, castillo2012needles}, which combine a point mass at zero with a diffuse continuous component.
However, spike-and-slab priors pose substantial computational challenges, as posterior inference involves a combinatorial model space with discrete inclusion indicators and prevents efficient block updates of model parameters.
An alternative class comprises continuous global-local shrinkage priors \citep{polson2010shrink}, which adaptively shrink small signals while preserving large ones.
Among these, the horseshoe prior \citep{carvalho2010horseshoe} has emerged as a popular and effective choice, enabling efficient block sampling strategies. In the context of Bayesian neural networks, \citep{ghosh2019model} demonstrated that horseshoe priors can induce sparsity at the level of network nodes, enabling automatic model selection and improved interpretability. Motivated by these results, we employ horseshoe priors over the parameters of generative neural networks to encourage sparse representation.

Let $\mathbf{\theta}$ denote the vector of all weights and biases in the network. The horseshoe prior is specified hierarchically as
\begin{eqnarray}
\label{ChapHSeq:hs_prior}
& \theta_j \mid \lambda_j, \tau \sim \mathcal{N}\!\left(0, \lambda_j^2 \tau^2 \right)\ \forall \ j=1, \ldots p, \\\nonumber
&\lambda_j \sim C^{+}(0, b_0) \ \forall \ j=1, \ldots p, \qquad
\tau \sim C^{+}(0, b_g),
\end{eqnarray}
where $\mathbf{\theta}=(\theta_1, \ldots, \theta_p)$. Here, $\lambda_j$ are parameter-specific local shrinkage parameters for all $j=1,\ldots,p$; and $\tau$ is a global shrinkage parameter that controls the overall degree of shrinkage. $C^{+}$ denotes the half-Cauchy distribution, and $b_0$ and $b_g$ are hyperparameters governing the prior distributions of the local and global shrinkage parameters, respectively.

\subsection{Blocked Gibbs sampler}
Combining Equation~\eqref{ChapHSeq:gen_post} and Equation~\eqref{ChapHSeq:hs_prior}, we can define the generalized posterior over the hierarchical model parameters as following, 
\begin{eqnarray}
\label{ChapHSeq:hier_post}
\mathbf{\theta} \mid \mathcal{D}_n, \left(\lambda_j:j=1,\ldots,p\right), \tau
\sim \pi(\mathbf{\theta})\exp\!\left\{-\omega\sum_{i=1}^n S(P_\mathbf{\theta},y_i)\right\}\prod_{j=1}^p\mathcal{N}\!\left(\theta_j | 0, \lambda_j^2 \tau^2 \right), \\\nonumber
\end{eqnarray}
where $\lambda_j \sim C^{+}(0, b_0) \ \forall \ j=1, \ldots p$ and $\tau \sim C^{+}(0, b_g)$.
Here, exact block sampling would require conjugacy between prior and likelihood, which breaks down for parameters $\theta$ as we have replaced log-likelihood with proper scoring rules. The posterior on the hyper-parameters, however, still remains conjugate.
Thus, we propose an approximate sampling scheme based on stochastic gradient Monte Carlo Markov chain (SGMCMC) \citep{nemeth2021stochastic} for $\mathbf{\theta}$ conditional on hyperparameters and an exact conditional update for the hyperparameters conditional on $\mathbf{\theta}$. This is an approximate blocked SGMCMC within the Gibbs algorithm.

When $\mathbf{\theta} = (\theta_j)_j$ are the network parameters and $\alpha = ((\lambda_j^2)_j, \tau^2)$ are the hyperparameters, to perform the Gibbs sampling we update $\mathbf{\theta}$ and $\alpha$ in an iterative manner using the following update equations. 
\paragraph{Block 1: Update parameter $\mathbf{\theta}$ conditional on hyperparameters $\alpha$ and data $\mathcal{D}_n$}
\begin{align}\label{ChapHSeqn: block1 param update}
\pi(\mathbf{\theta} \mid \alpha, \mathcal{D}_n) 
&\propto\exp \Biggl( -\sum_{i=1}^{n}S_E(P_{\theta}, y_{i})\Biggl) { \prod_{j=1}^{p} \frac{1}{\sqrt{2 \pi \lambda_{j}^2 \tau^2}} \exp\left(-\frac{1}{2 \lambda_{j}^2 \tau^2} \theta_{j}^{2}\right)}\\\nonumber
&\propto \exp \left(-\sum_{i=1}^{n}S_E\left(P_{\theta}, y_{i}\right) -\sum_{j=1}^{p}\frac{1}{2 \lambda_{j}^2 \tau^2} \theta_{j}^{2}\right).
\end{align}
We update $\mathbf{\theta}$ by using a variant of stochastic gradient Monte Carlo Markov chain algorithm (SGMCMC), specifically preconditioned adjusted stochastic gradient Langevin dynamics (prec-adjSGLD) described in Section~\ref{ChapHSsec:adjSGLD}
\paragraph{Block 2: Update hyperparameters conditional on $\mathbf{\theta}$} 
Half-Cauchy priors on $\lambda_j$ and $\tau$ break conjugacy. However, a scale-mixture representation of the Half-Cauchy prior proposed by \citep{makalic2015simple}, can be used to arrive at the following update equations using the auxiliary variables $ \nu_0, \nu_1, \ldots \nu_p \sim \mathcal{IG}(1/2, 1)$, 
\begin{align}\label{ChapHSeqn: block2 hyperparameters}
    \lambda_{j}^2| \cdot & \sim \mathcal{IG}(1, \frac{\theta^2_{j}}{2 \tau^2_{}} + \frac{1}{\nu_{j}})\\\nonumber
    \tau^2| \cdot & \sim \mathcal{IG}(\frac{p + 1}{2}, \sum_{j} \frac{\theta^2_{j}}{2\lambda^2_{j}} + \frac{1}{\nu_{0}})\\\nonumber
    \nu_{j} |\cdot & \sim \mathcal{IG}(1, \frac{1}{b^2_0} + \frac{1}{\lambda^2_{j}})\\\nonumber
    \nu_0|\cdot & \sim \mathcal{IG}(1, \frac{1}{b^2_g} + \frac{1}{\tau^2}),
\end{align}
for all $j = 1, 2, \ldots p$, where $\mathcal{IG} (a, b)$ is the inverse-gamma distribution with shape parameter $a >0$ and scale parameter $b>0$. Hence, we can update hyperparameters $\alpha$ exactly, using the above update equations.


\subsection{Prec-adjSGLD to update Block 1}
\label{ChapHSsec:adjSGLD}

We propose a variant of the SGMCMC algorithm to update the model parameters in Block 1. 
SGMCMC methods are based on continuous-time diffusion processes using gradients of the negative log target distribution (or the potential function $U(\mathbf{\theta})$); whose stationary distribution matches the target distribution. In our case, $U(\mathbf{\theta})=\sum_{i=1}^{n}S_E\left(P_{\theta}, y_{i}\right) + \sum_{j=1}^{p}\frac{1}{2 \lambda_{j}^2 \tau^2} \theta_{j}^{2}$, for which the gradient is not analytically tractable. 

But we can compute an unbiased estimate of the gradient, by first noticing that $S_E(P_\mathbf{\theta}, y) = \displaystyle \mathop{\mathbb{E}}_{X, X^{\prime} \sim P_\theta} f\left(X, X^{\prime}, y\right)$  for a smooth differentiable function $f$. Thus, the scoring rule is defined through an expectation over (possibly multiple) samples from $P_\mathbf{\theta}$. 
Hence the gradient of the scoring rule reduces to,
\begin{align*}
\nabla_\mathbf{\theta} S_E(P_\mathbf{\theta}, y) &= \nabla_\mathbf{\theta} \displaystyle \mathop{\mathbb{E}}_{X, X^{\prime} \sim P_\mathbf{\theta}} f\left(X, X^{\prime}, y\right) \\
&= \nabla_\mathbf{\theta} \displaystyle \mathop{\mathbb{E}}_{Z, Z^{\prime} \sim q} f\left(G_{\mathbf{\theta}}(Z), G_{\mathbf{\theta}}(Z^{\prime}), y\right)\\
&=  \displaystyle \mathop{\mathbb{E}}_{Z, Z^{\prime} \sim q} \nabla_\theta f\left(G_{\mathbf{\theta}}(Z), G_{\mathbf{\theta}}(Z^{\prime}), y\right),
\end{align*}
where $q$ stands for an auxiliary distribution independent of $\theta$. In the latter equality, the exchange between expectation and gradient is not a trivial step.
Luckily, Theorem 5 in \citep{binkowski2018demystifying} proved the above step to be valid almost surely with respect to a measure on $\mathbf{\theta}$, under mild conditions on $G_{\mathbf{\theta}}$ (such conditions are satisfied if the function is differentiable, which holds for all the maps defined by neural networks considered in this work). Based on this, we can unbiasedly estimate the gradient of the scoring rule as follow:
\begin{eqnarray}
\label{ChapHSeq:uest_gradSR}
&&\widehat{\nabla_\theta} S_E(P_\theta, y) =\frac{1}{m(m-1)} \sum_{\substack{i, j=1 \\ i \neq j}}^m 
\nabla_\theta f\left(G_{\mathbf{\theta}}(Z_i), G_{\mathbf{\theta}}(Z_j^{\prime}), y\right)
\end{eqnarray}
where $Z_i \perp \!\!\! \perp Z^{\prime}_j \sim q$. In practice, this can be easily obtained by implementing the function $G_\theta$ using automatic-differentiation libraries \citep{pytorch}. 

The bias due to this approximate stochastic gradient of $U(\mathbf{\theta})$ in SGMCMC algorithms can be mitigated by incorporating momentum ($\boldsymbol p$) and friction ($\boldsymbol \xi$) terms, resulting in the \emph{adjusted stochastic gradient Langevin dynamics} (adjSGLD) \citep{leimkuhler2016adaptive}.  
adjSGLD operates in an extended state space consisting of the model parameters, and the auxiliary momentum and friction variables. This has been successfully applied in Bayesian inference \citep{ding2014bayesian} as well. 

To improve the efficiency of SGMCMC in the high-dimensional parameter space of $\mathbf{\theta}$, we can use preconditioning strategies \citep{girolami2011riemann, chen2016bridging} using the squared gradient of $U( {\theta})$, similar to adaptive optimizers like RMSprop \citep{tieleman2012lecture} and Adam \citep{adam2014method}.
Here we use a preconditioned version of the adjSGLD scheme (\emph{prec-adjSGLD}) to sample $\mathbf{\theta}$, for which the stochastic differential equation (SDE) defining the diffusion becomes,
\begin{equation}
        \label{ChapHSeq:adSGLD}
        \left\{ \begin{aligned}
        d \boldsymbol \theta &=H(\boldsymbol \theta)\boldsymbol p d t \\
        d \boldsymbol p &=(-H(\boldsymbol \theta) \nabla_ {\theta} U(\boldsymbol {\theta}) - \boldsymbol \xi \boldsymbol p)d t+ (2H(\boldsymbol \theta))^{1/2} dW_t \\
        d \boldsymbol \xi &= \left(\frac{1}{p} \boldsymbol p_{}^{\top} \boldsymbol p_{}-1\right) d t
        \end{aligned} \right.
        \end{equation}
where $W$ denotes a standard Brownian motion and $H(\boldsymbol{\theta})$ encodes geometric information of the potential energy $U(\mathbf{\theta})$, called the Riemannian metric \citep{girolami2011riemann}, which is commonly defined by the Fisher information matrix. Theoretical properties and convergence of adSGLD and prec-adjSGLD algorithm have been studied in \citep{ leimkuhler2016adaptive, leimkuhler2020hypocoercivity} and \citep{ding2014bayesian} correspondingly.

The other source of error in SGMCMC arises due to the time discretization of the above SDE. For example, when the first order Euler–Maruyama discretization scheme is used.    
To further reduce these numerical errors in SDE integration, we use a symmetric splitting scheme (SSS) \citep{chen2016bridging}, described in Appendix~\ref{ChapHSsec:splitscheme}.  
This 2nd-order integrator provides higher accuracy with minimal extra computation, and we arrive at Algorithm~\ref{ChapHSalg:prec-adjSGLD-santa} presented below.

\begin{algorithm}[H]
\caption{Preconditioned adjusted stochastic gradient Langevin dynamics } 
\label{ChapHSalg:prec-adjSGLD-santa}


\KwIn{Learning rate $\eta$, parameters $(\lambda_0 = 0.99, \lambda_1 = 10^{-6})$, noise $\{ \boldsymbol{\zeta}_t \sim \mathcal{N}(\mathbf{0}, \mathbf{I}_p) \}$}
\KwInit{$\boldsymbol{\theta}_0$, $\mathbf{u}_0 = \sqrt{\eta} \cdot \mathcal{N}(0, \mathbf{I}_p)$, $\boldsymbol{a}_0 = \sqrt{\eta}$, $\mathbf{v}_0 = \mathbf{0}$}

\For{$t = 1, 2, \ldots $}{
    Compute stochastic gradient: $\tilde{\mathbf{f}}_t \gets \nabla_{\boldsymbol{\theta}} \tilde{U}(\boldsymbol{\theta}_{t-1})$\

    $\mathbf{v}_t \gets \lambda_0 \mathbf{v}_{t-1} + \frac{1 - \lambda_0}{n^2} \tilde{\mathbf{f}}_t \odot \tilde{\mathbf{f}}_t$\

    $\mathbf{g}_t \gets 1 \oslash \sqrt{\lambda_1 + \sqrt{\mathbf{v}_t}}$\

    $\boldsymbol{\theta}_t \gets \boldsymbol{\theta}_{t-1} + \frac{1}{2} \mathbf{g}_t \odot \mathbf{u}_{t-1}$\

    $\boldsymbol{a}_t \gets \boldsymbol{a}_{t-1} + \frac{1}{2} \left( \mathbf{u}_{t-1} \odot \mathbf{u}_{t-1} - \eta \right)$\

    $\mathbf{u}_t \gets \exp(-\boldsymbol{a}_t / 2) \odot \mathbf{u}_{t-1}$\

    $\mathbf{u}_t \gets \mathbf{u}_t - \eta \cdot \mathbf{g}_t \odot \tilde{\mathbf{f}}_t + \sqrt{2 \cdot \mathbf{g}_{t-1} \eta^{3/2}} \odot \boldsymbol{\zeta}_t$\

    $\mathbf{u}_t \gets \exp(-\boldsymbol{a}_t / 2) \odot \mathbf{u}_t$\

    $\boldsymbol{a}_t \gets \boldsymbol{a}_t + \frac{1}{2} \left( \mathbf{u}_t \odot \mathbf{u}_t - \eta \right)$\

    $\boldsymbol{\theta}_t \gets \boldsymbol{\theta}_t + \frac{1}{2} \mathbf{g}_t \odot \mathbf{u}_t$
}

\textbf{end}

\end{algorithm}

\section{Simulation studies}\label{ChapHSsec:simulations}

To evaluate our method, we first consider a sparse linear regression task with a small number of observations and varying numbers of regressors.
In this setting, we assume Gaussian noise, which allows access to the exact likelihood.
We compare the performance of the (likelihood-based) exact Gibbs sampler with our proposed (likelihood-free) approximate sampler under the horseshoe prior.
Finally, we consider a Markov decision process, called Hopper. Here, the simulator model is a conditional generative model and we analyze the effect of different priors on the model parameters compared to the horseshoe prior.

\subsection{Bayesian sparse high-dimensional linear regression}

Let us consider a high-dimensional linear regression problem,
\begin{equation}
\mathbf{y}_{n\times 1} = \mathbf{W_{n\times p}}\mathbf{\theta}_{p\times 1} + \epsilon_{1\times n},
\end{equation}
where $\epsilon \sim \mathcal{N}(\mathbf{0}, \sigma^2\mathbf{I}_n)$, $\mathbf{I}_n$ is $n\times n$ identity matrix and $p\gg n$. Inference for $\mathbf{\theta}$ under sparsity has been widely addressed using penalized estimation methods. Classical examples include ridge regression \citep{hoerl1970ridge}, LASSO \citep{tibshirani1996regression}, and elastic net \citep{zou2005regularization}, among many other regularization approaches. Similarly, Bayesian inference for $\mathbf{\theta}$ can be performed by using a shrinkage-inducing prior, like a horseshoe prior. Although the true likelihood function is available in this case, the traditionally used Gibbs sampler is prohibitively expensive, and approximations to sample from the posterior is an active research area \citep{bhadra2019lasso, johndrow2020scalable}.

Considering sparse linear regression as an implicit generative model, we demonstrate inference on $\theta$ by sampling from a scoring-rule posterior using our proposed methodology. It is scalable and achieves superior predictive performance. In our experiments, we fix the sample size at $n = 50$ and consider three settings with $p = 500, 1000,$ and $3000$.

Our experimental setup follows that of \citet{johndrow2020scalable}. We set $\sigma^2 = 4$, and generate $W$ from a standard Gaussian distribution. We define the true sparse parameter vector $\theta$ as
\[
\theta_j =
\begin{cases}
2^{-(j/4 - 9/4)}, & \text{if } j < 24, \\
0, & \text{otherwise}.
\end{cases}
\]

\begin{figure}[htbp!]
    \centering
    \includegraphics[width=0.99\linewidth]{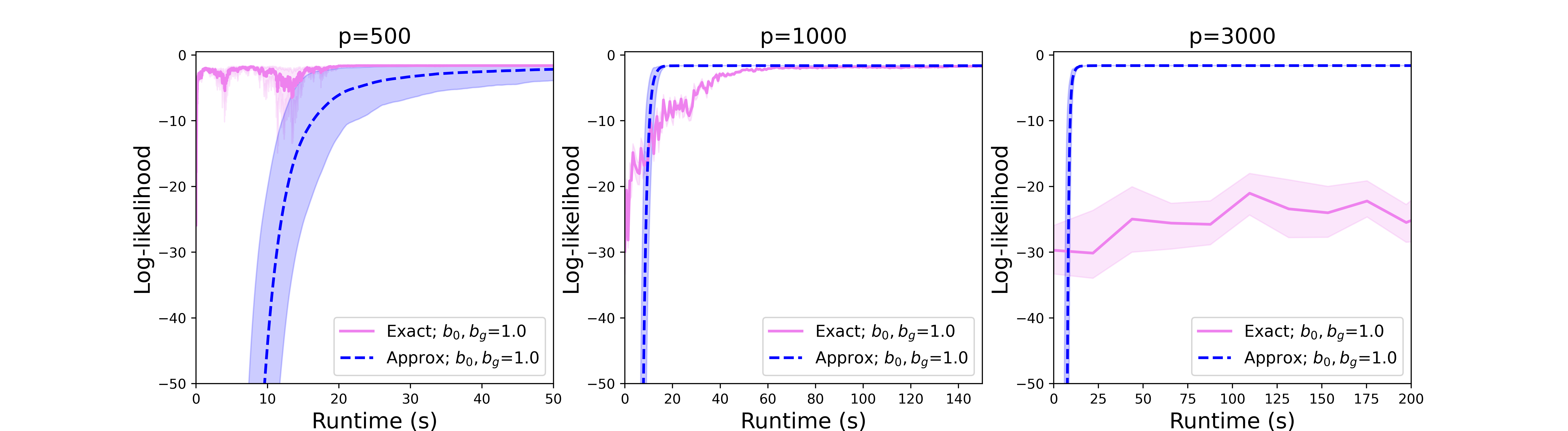}
    \caption{Exact versus approximate sampler for sparse linear regression. Log likelihood of the samples obtained from the exact (in pink) and the approximate (in blue) sampler versus runtime plot. The solid line represents the mean performance over $10$ independent runs, and the $90\ \%$ confidence intervals are shown in a lighter shade.  }
    \label{ChapHSfig:lreg-llhd}
\end{figure}

We compare the proposed sampler with an exact Gibbs sampler, which is feasible in this setting since the likelihood is available in closed form. For both methods, we use $b_0=b_g=1$ to define the horseshoe prior. The sensitivity to these hyperparameters is investigated in Appendix~\ref{ChapHSappendix:lreg_bvlaues_llhd}, where we also consider $(b_0,b_g)=(10,10)$ and $(50,50)$. We set the learning rate to $\eta = 10^{-8}$ in Algorithm~\ref{ChapHSalg:prec-adjSGLD-santa} and we use $m=10$ simulations to estimate the gradient in Equation~\eqref{ChapHSeq:uest_gradSR} for the approximate sampler.
Both the samplers are run under matched time budgets. In Figure~\ref{ChapHSfig:lreg-llhd}, we report the log likelihood of the posterior samples.

 Figure~\ref{ChapHSfig:lreg-llhd} illustrates that our method exhibits improved sample efficiency as the dimensionality $p$ increases, while the exact Gibbs sampler struggles in higher-dimensional settings. All experiments are repeated across ten independently generated datasets, and the results are averaged.

For comparison, we also implement ridge regression, LASSO, and elastic net for the linear regression model. For these methods, we evaluate the log-likelihood at the estimated parameter. For the Bayesian approach with the horseshoe prior, we compute the mean log-likelihood over posterior samples obtained from both the exact and the proposed approximate samplers. To ensure a fair comparison, we discard the initial samples corresponding to the first 25, 75, and 100 seconds of runtime for $p=500$, $1000$, and $3000$, respectively. The mean log-likelihood is then computed using the remaining samples. Table~\ref{ChapHStab:linreg-llhd} reports the results averaged over the 10 simulated datasets.

\begin{table}[htbp!]
\centering
\begin{tabular}{lccc}
\toprule
Method & $p=500$ & $p=1000$ & $p=3000$ \\
\midrule
Ridge                          & $-30.79\ (45.52)$  & $-36.62\ (65.32)$  & $-127.20\ (165.12)$ \\
LASSO                          & $-34.70\ (36.70)$  & $-98.18\ (80.73)$  & $-111.54\ (116.90)$ \\
Elastic Net                    & $-34.34\ (37.00)$  & $-88.61\ (84.56)$  & $-92.38\ (124.10)$ \\
Horseshoe (exact Gibbs)        & $\mathbf{-1.616}\ (0.001)$  & $-1.830\ (0.032)$  & $-23.290\ (1.169)$ \\
Horseshoe (approximate, ours)  & $-2.805\ (1.486)$  & $\mathbf{-1.626}\ (0.035)$  & $\mathbf{-1.612}\ (2\times 10^{-5})$ \\
\bottomrule
\end{tabular}
\caption{Mean log-likelihood (standard deviation ) across 10 simulated datasets for the linear regression example. For the first three methods, the log-likelihood at the estimated parameter value is computed. For the Bayesian methods, the mean log-likelihood is computed over posterior samples after discarding the initial burn-in period. Higher values indicate better predictive performance.}
\label{ChapHStab:linreg-llhd}
\end{table}

The Bayesian methods consistently outperform the three frequentist regularization methods in this example. Moreover, the approximate sampler performs better in larger dimensions, consistent with the trends observed in Figure~\ref{ChapHSfig:lreg-llhd}.

\begin{table}[h!]
\centering
\begin{tabular}{l c c c}
\hline
 & $p=500$ & $p=1000$ & $p=3000$ \\
\hline
Exact Sampler 
& 0.0987 (0.0019) 
& 0.6736 (0.0218) 
& 21.9139 (1.8124) \\
\hline
Approximate Sampler 
& 0.0309 (0.0025) 
& 0.0310 (0.0015) 
& 0.0345 (0.0005) \\
\hline
\end{tabular}
\caption{Mean runtime (in seconds) per iteration across 10 seeds for exact and approximate samplers for linear regression. Standard deviation over the 10 runs are shown in parentheses.}
\label{ChapHStab:sampler_time-lreg-hs}
\end{table}

Next, we report the runtime comparison of each iteration of the exact and approximate samplers in Table~\ref{ChapHStab:sampler_time-lreg-hs}. The runtime of the exact sampler blows up with dimension, whereas the proposed approximate sampler does not have that trend.

\subsection{Neural generative Markov processes: Hopper}

Next, we consider a model-based reinforcement learning (RL) task. In model-based RL, the environment dynamics are first learned from interactions with the environment. The learned model is then used to train a sample-efficient policy. Therefore, improving model learning with fewer interactions can directly benefit the final policy performance. This motivates the use of shrinkage priors, which are known to perform well in data-scarce settings.

We consider the `Hopper' environment from the OpenAI Gymnasium package \citep{kwiatkowski2024gymnasiumstandardinterfacereinforcement}. The task is to move a two-dimensional, single-legged structure forward by applying forces to three joints connecting its four body parts. The action space consists of torques applied to the three joints, with values ranging from $-1$ Nm to $1$ Nm. The state is represented by a $12$-dimensional real-valued vector containing joint angles, angular velocities, and the positions and velocities of the body parts. The reward function encourages forward movement while penalizing large torques that may destabilize the system.

We generate a trajectory of state-action-next state from the above system using a random policy. We wish to model this Markov process by a deep generative model following the setup of \citet{roy2024generalized}. Here, we consider the following conditional generative model,
\begin{equation*}
    f_{\theta}(s_t, a_t, z) = s_{t+1}
\end{equation*}
where $z$ is a Gaussian noise, $a_t$ and $s_t$ are action and state at time $t$ and $f_\theta$ is fully connected feedforward neural network with four hidden layers of $100$ neurons each and swish activation function, which amounts to a total number of parameters approximately 33,000.

To infer the model parameters ${\theta}$, we consider a scoring-rule prequential posterior \citep{Sinha_Roy_2026} under three different priors: a horseshoe prior with hyperparameters $b_0=b_g=1$, an independent standard Gaussian prior, and an independent Student-$t$ prior with 10 degrees of freedom on each network parameter. We focus on the data-scarce regime with $n=10,20,50,$ and $100$ observations, in contrast to the conventional setting where thousands of observations are used for training. For sampling, we set the learning rate to $\eta=10^{-12}$ in Algorithm~\ref{ChapHSalg:prec-adjSGLD-santa} and use $m=10$ simulations to estimate the gradient in Equation~\eqref{ChapHSeq:uest_gradSR}. Each sampler is run for $10000$ iterations, and the last $1000$ samples are retained to evaluate predictive performance.

\begin{figure*}[htbp!]
    \centering
    \includegraphics[width=0.95\linewidth]{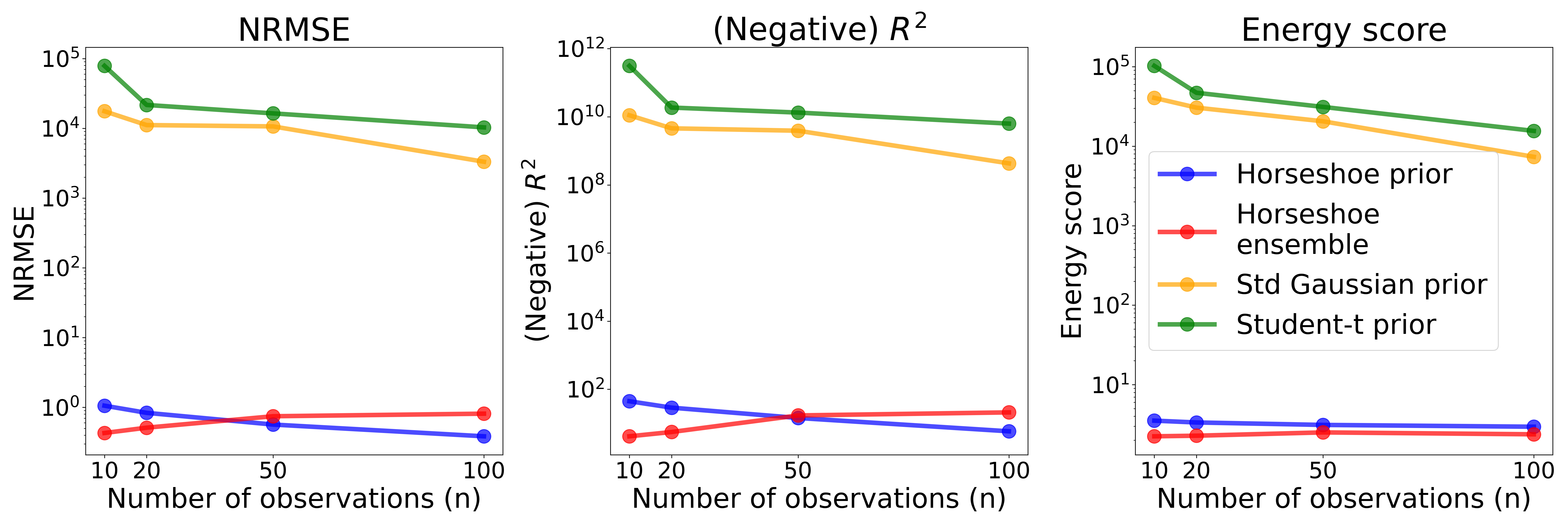}
    \caption{Predictive performance on the \emph{Hopper} environment. We report the normalized root mean squared error (NRMSE), coefficient of determination $R^2$, and average energy score on a test set of length $1000$. Results are shown for the horseshoe prior (blue), standard Gaussian prior (yellow), Student-$t$ prior with 10 degrees of freedom (green), and an ensemble model initialized from the horseshoe prior and optimized using Adam (red). For visual clarity, the negative adjusted $R^2$ is plotted.}
    \label{ChapHSfig:hopper_perf}
\end{figure*}

Figure~\ref{ChapHSfig:hopper_perf} reports the NRMSE, coefficient of determination $R^2$, and average energy score on an independent test set of $1000$ observations. We also consider an ensemble model in which the network parameters are initialized from the horseshoe prior and optimized using Adam. The objective function is estimated using $m=10$ simulations. The horseshoe prior consistently outperforms the Gaussian and Student-$t$ priors, especially when the number of training observations is small. The ensemble model also performs well for small values of $n$, suggesting that the horseshoe prior helps recover the sparse structure of the dynamics model. In contrast, the Gaussian and Student-$t$ priors produce substantially larger prediction errors, indicating poorer calibration when only fewer observations are available.

\begin{table}[htbp!]
\centering
\begin{tabular}{lccc}
\hline
$n$ & NRMSE & $R^2$ & Energy score \\
\hline
50   & 16998.45 (3058.44) & -2.31e10 (8.50e9) & 78153.29 (16198.17) \\
100  & 17301.19 (2740.90) & -2.58e10 (6.68e9) & 80490.82 (13381.43) \\
500  & 16442.49 (3102.66) & -2.74e10 (1.05e10) & 63382.63 (9684.82) \\
1000 & 18561.19 (2940.66) & -3.27e10 (1.11e10) & 70151.97 (12845.99) \\
\hline
\end{tabular}
\caption{Predictive performance of the posterior samples under a standard Gaussian prior for different training set sizes $(n)$. The Markov chain is run for $10000$ iterations, and the last $1000$ samples are used to compute the predictive metrics. Results are reported as mean (standard deviation) over five independent runs.}
\label{ChapHStab:Hopper-gaussian calibration}
\end{table}

\begin{table}[htbp!]
\centering
\begin{tabular}{lccc}
\hline
$n$ & NRMSE & $R^2$ & Energy score \\
\hline
50   & 0.212 (0.070) & -3.070 (3.260) & 3.001 (0.079) \\
100  & 0.215 (0.074) & -3.120 (3.750) & 2.984 (0.065) \\
500  & 0.152 (0.073) & -1.550 (3.200) & 2.934 (0.046) \\
1000 & 0.115 (0.055) & -0.470 (1.890) & 2.914 (0.031) \\
\hline
\end{tabular}
\caption{Predictive performance of the posterior samples under the horseshoe prior with $b_0=b_g=100$. The Markov chain is run for $10000$ iterations, and the last $1000$ samples are used to compute the predictive metrics. Results are reported as mean (standard deviation) over five independent runs.}
\label{ChapHStab:Hopper-b1}
\end{table}

We next study the sensitivity of the horseshoe hyperparameters for additional cases when $n=500$ and $1000$. We consider five independently generated trajectories and compare the standard Gaussian prior with horseshoe priors using five different choices of $(b_0,b_g)$. Table~\ref{ChapHStab:Hopper-gaussian calibration} reports the results for the Gaussian prior, while Table~\ref{ChapHStab:Hopper-b1} reports the results for the horseshoe prior with $b_0=b_g=1$. The remaining results are provided in Appendix~\ref{ChapHSappendix:hopper_hs}. Overall, the predictive performance is mostly insensitive to the choice of horseshoe hyperparameters when $b_0, b_g \geq 1$. Finally, Table~\ref{ChapHStab:hopper-runtime} reports the average runtime per iteration for the Gaussian and horseshoe priors.

\begin{table}[ht]
\centering
\begin{tabular}{lcccc}
\toprule
Method & $n=50$ & $n=100$ & $n=500$ & $n=1000$ \\
\midrule
Gaussian prior  & $0.062\ (0.002)$ & $0.105\ (0.006)$ & $0.487\ (0.022)$ & $1.077\ (0.006)$ \\
Horseshoe prior & $0.118\ (0.016)$ & $0.153\ (0.012)$ & $0.517\ (0.028)$ & $1.208\ (0.050)$ \\
\bottomrule
\end{tabular}
\caption{Average runtime (seconds) per iteration for the Gaussian and horseshoe priors. Values are reported as mean (standard deviation) over repeated runs.}
\label{ChapHStab:hopper-runtime}
\end{table}

Table~\ref{ChapHStab:hopper-runtime} reports the runtime of the approximate sampler using the Gaussian and horseshoe priors. The horseshoe prior requires slightly more computational time due to the additional componentwise updates of the shrinkage parameters. However, this increase in runtime is small compared to the substantial predictive improvements obtained. These results highlight the advantage of using adaptive shrinkage priors, such as the horseshoe prior, in severely data-poor settings.



\section{Discussion and conclusion}\label{ChapHSsec:conclusion}

In recent years, neural implicit generative models have been rapidly adopted across many application domains. This work extends their applicability to settings with limited training data, which may arise due to data collection constraints or high acquisition costs. We achieve this by incorporating shrinkage priors within the scoring-rule posterior framework.

First, we demonstrate the strong predictive performance of the proposed methodology on a high-dimensional sparse linear regression. Next, we implement our method on a neural generative model, motivated by applications in meteorology and reinforcement learning. Beyond enabling scalable training of larger neural architectures with fewer observations, our approach also helps identify sparse structures within generative neural networks due to the shrinkage prior. Further investigation in this direction may lead to more meaningful and potentially interpretable neural network architectures for specific applications.

In this work, we fix the scaling parameter of the scoring-rule posterior, $\omega$. Future work could investigate the use of multivariate scaling parameters or data-driven approaches, such as cross-validation, to select $\omega$ and obtain more informative scoring rules. Another interesting direction is to develop model selection approaches for implicit generative models, or for combinations of models and scoring rules viewed as approximate likelihoods, through comparisons of their marginal scoring-rule posteriors.

\section{Acknowledgment}
We thank Sherman Khoo for his initial investigations into the horseshoe prior and for many fruitful discussions.


\bibliography{references}

@article{nemeth2021stochastic,
  title={Stochastic gradient {{M}arkov chain {M}onte {C}arlo}},
  author={Nemeth, Christopher and Fearnhead, Paul},
  journal={Journal of the American Statistical Association},
  volume={116},
  number={533},
  pages={433--450},
  year={2021},
  publisher={Taylor \& Francis}
}

@article{pacchiardi2024generalized,
  title={Generalized {{B}ayesian} likelihood-free inference},
  author={Pacchiardi, Lorenzo and Khoo, Sherman and Dutta, Ritabrata},
  journal={Electronic Journal of Statistics},
  volume={18},
  number={2},
  pages={3628--3686},
  year={2024},
  publisher={The Institute of Mathematical Statistics and the Bernoulli Society}
}

@article{mirza2014conditional,
  title={Conditional generative adversarial nets},
  author={Mirza, Mehdi and Osindero, Simon},
  journal={arXiv preprint arXiv:1411.1784},
  year={2014}
}

@article{pacchiardi2024probabilistic,
  title={Probabilistic Forecasting with Generative Networks via Scoring Rule Minimization},
  author={Pacchiardi, Lorenzo and Adewoyin, Rilwan and Dueben, Peter and Dutta, Ritabrata},
  journal={Journal of Machine Learning Research},
  year={2024},
  volume={25},
  pages={1--64}
}

@phdthesis{pacchiardi2022statistical,
  title={Statistical inference in generative models using scoring rules},
  author={Pacchiardi, Lorenzo},
  year={2022},
  school={University of Oxford}
}

@article{gneiting2007strictly,
  title={Strictly proper scoring rules, prediction, and estimation},
  author={Gneiting, Tilmann and Raftery, Adrian E},
  journal={Journal of the American statistical Association},
  volume={102},
  number={477},
  pages={359--378},
  year={2007},
  publisher={Taylor \& Francis}
}

@article{ding2014bayesian,
  title={{B}ayesian sampling using stochastic gradient thermostats},
  author={Ding, Nan and Fang, Youhan and Babbush, Ryan and Chen, Changyou and Skeel, Robert D and Neven, Hartmut},
  journal={Advances in neural information processing systems},
  volume={27},
  year={2014}
}

@article{leimkuhler2016adaptive,
  title={Adaptive thermostats for noisy gradient systems},
  author={Leimkuhler, Benedict and Shang, Xiaocheng},
  journal={SIAM Journal on Scientific Computing},
  volume={38},
  number={2},
  pages={A712--A736},
  year={2016},
  publisher={SIAM}
}

@article{leimkuhler2020hypocoercivity,
  title={Hypocoercivity properties of adaptive {Langevin} dynamics},
  author={Leimkuhler, Benedict and Sachs, Matthias and Stoltz, Gabriel},
  journal={SIAM Journal on Applied Mathematics},
  volume={80},
  number={3},
  pages={1197--1222},
  year={2020},
  publisher={SIAM}
}

@inproceedings{binkowski2018demystifying,
  title={Demystifying {MMD GAN}s},
  author={Bi{\'n}kowski, Miko{\l}aj and Sutherland, Danica J and Arbel, Michael and Gretton, Arthur},
  booktitle={International Conference on Learning Representations},
  year={2018}
}

@article{girolami2011riemann,
  title={{Riemann} manifold {Langevin and Hamiltonian {M}onte {C}arlo} methods},
  author={Girolami, Mark and Calderhead, Ben},
  journal={Journal of the Royal Statistical Society Series B: Statistical Methodology},
  volume={73},
  number={2},
  pages={123--214},
  year={2011},
  publisher={Oxford University Press}
}

@inproceedings{kwiatkowski2024gymnasiumstandardinterfacereinforcement,
      title={Gymnasium: A Standard Interface for Reinforcement Learning Environments},
      author={Ariel Kwiatkowski and Mark Towers and Jordan Terry and John U. Balis and Gianluca De Cola and Tristan Deleu and Manuel Goulão and Andreas Kallinteris and Markus Krimmel and Arjun KG and Rodrigo Perez-Vicente and Andrea Pierré and Sander Schulhoff and Jun Jet Tai and Hannah Tan and Omar G. Younis},
      booktitle={Advances in Neural Information Processing Systems 38 (NeurIPS 2025), Datasets and Benchmarks Track},
      year={2025},
      note={Spotlight}
}

@inproceedings{chen2016bridging,
  title={Bridging the gap between stochastic gradient {MCMC} and stochastic optimization},
  author={Chen, Changyou and Carlson, David and Gan, Zhe and Li, Chunyuan and Carin, Lawrence},
  booktitle={Artificial Intelligence and Statistics},
  pages={1051--1060},
  year={2016},
  organization={PMLR}
}

@article{bissiri2016general,
  title={A general framework for updating belief distributions},
  author={Bissiri, Pier Giovanni and Holmes, Chris C and Walker, Stephen G},
  journal={Journal of the Royal Statistical Society: Series B (Statistical Methodology)},
  volume={78},
  number={5},
  pages={1103--1130},
  year={2016},
  publisher={Wiley Online Library}
}

@article{roy2024generalized,
  title={Generalized {B}ayesian deep reinforcement learning},
  author={Sinha-Roy, Shreya and Everitt, Richard G and Robert, Christian P and Dutta, Ritabrata},
  journal={arXiv preprint arXiv:2412.11743},
  year={2024}
}

@article{jewson2018principles,
  title={Principles of {B}ayesian inference using general divergence criteria},
  author={Jewson, Jack and Smith, Jim Q and Holmes, Chris},
  journal={Entropy},
  volume={20},
  number={6},
  pages={442},
  year={2018},
  publisher={MDPI}
}

@article{knoblauch2022optimization,
  title={An optimization-centric view on Bayes' rule: Reviewing and generalizing variational inference},
  author={Knoblauch, Jeremias and Jewson, Jack and Damoulas, Theodoros},
  journal={Journal of Machine Learning Research},
  volume={23},
  number={132},
  pages={1--109},
  year={2022}
}

@article{price2024gencastdiffusionbasedensembleforecasting,
      title={Probabilistic weather forecasting with machine learning},
      author={Ilan Price and Alvaro Sanchez-Gonzalez and Ferran Alet and Tom R. Andersson and Andrew El-Kadi and Dominic Masters and Timo Ewalds and Jacklynn Stott and Shakir Mohamed and Peter Battaglia and Remi Lam and Matthew Willson},
      journal={Nature},
      volume={637},
      number={8044},
      pages={84--90},
      year={2025},
      publisher={Nature Publishing Group},
      note={Preprint title: ``GenCast: Diffusion-based ensemble forecasting for medium-range weather.''}
}

@article{tibshirani1996regression,
  title={Regression shrinkage and selection via the {LASSO}},
  author={Tibshirani, Robert},
  journal={Journal of the Royal Statistical Society Series B: Statistical Methodology},
  volume={58},
  number={1},
  pages={267--288},
  year={1996},
  publisher={Oxford University Press}
}

@article{zou2005regularization,
  title={Regularization and variable selection via the elastic net},
  author={Zou, Hui and Hastie, Trevor},
  journal={Journal of the Royal Statistical Society Series B: Statistical Methodology},
  volume={67},
  number={2},
  pages={301--320},
  year={2005},
  publisher={Oxford University Press}
}

@InCollection{pytorch,
  title		= {Py{T}orch: An Imperative Style, High-Performance Deep
		  Learning Library},
  author	= {Paszke, Adam and Gross, Sam and Massa, Francisco and
		  Lerer, Adam and Bradbury, James and Chanan, Gregory and
		  Killeen, Trevor and Lin, Zeming and Gimelshein, Natalia and
		  Antiga, Luca and Desmaison, Alban and Kopf, Andreas and
		  Yang, Edward and DeVito, Zachary and Raison, Martin and
		  Tejani, Alykhan and Chilamkurthy, Sasank and Steiner,
		  Benoit and Fang, Lu and Bai, Junjie and Chintala, Soumith},
  booktitle	= {Advances in Neural Information Processing Systems 32},
  editor	= {H. Wallach and H. Larochelle and A. Beygelzimer and F.
		  d'Alch\'{e}-Buc and E. Fox and R. Garnett},
  pages		= {8024--8035},
  year		= {2019},
  publisher	= {Curran Associates, Inc.}
}

@Article{pacchiardi2020score,
  author	= {Lorenzo Pacchiardi and Ritabrata Dutta},
  title		= {Score Matched Neural Exponential Families for
		  Likelihood-Free Inference},
  journal	= {Journal of Machine Learning Research},
  year		= {2022},
  volume	= {23},
  number	= {38},
  pages		= {1-71},}

@article{heng2024diffusion,
  title={Diffusion Schr{\"o}dinger bridges for {B}ayesian computation},
  author={Heng, Jeremy and De Bortoli, Valentin and Doucet, Arnaud},
  journal={Statistical Science},
  volume={39},
  number={1},
  pages={90--99},
  year={2024},
  publisher={Institute of Mathematical Statistics}
}

@article{tieleman2012lecture,
  title={Lecture 6.5-rmsprop: Divide the gradient by a running average of its recent magnitude},
  author={Tieleman, Tijmen},
  journal={COURSERA: Neural networks for machine learning},
  volume={4},
  number={2},
  pages={26},
  year={2012}
}

@inproceedings{fortuin2022bayesian,
  title = "{B}ayesian Neural Network Priors Revisited",
  author = "Vincent Fortuin and Adri{\`a} Garriga-Alonso and Sebastian W. Ober and Florian Wenzel and Gunnar Ratsch and Richard E Turner and Mark van der Wilk and Laurence Aitchison",
  year = "2022",
  booktitle = "International Conference on Learning Representations (ICLR)",
}

@article{dawid2007geometry,
  title={The geometry of proper scoring rules},
  author={Dawid, A Philip},
  journal={Annals of the Institute of Statistical Mathematics},
  volume={59},
  number={1},
  pages={77--93},
  year={2007},
  publisher={Springer}
}

@article{donoho2009observed,
  title={Observed universality of phase transitions in high-dimensional geometry, with implications for modern data analysis and signal processing},
  author={Donoho, David and Tanner, Jared},
  journal={Philosophical Transactions of the Royal Society A: Mathematical, Physical and Engineering Sciences},
  volume={367},
  number={1906},
  pages={4273--4293},
  year={2009},
  publisher={The Royal Society Publishing}
}

@Article{lyddon2019general,
  title		= {General {B}ayesian updating and the loss-likelihood
		  bootstrap},
  author	= {Lyddon, SP and Holmes, CC and Walker, SG},
  journal	= {Biometrika},
  volume	= {106},
  number	= {2},
  pages		= {465--478},
  year		= {2019},
  publisher	= {Oxford University Press}
}

@Article{syring2019calibrating,
  title		= {Calibrating general posterior credible regions},
  author	= {Syring, Nicholas and Martin, Ryan},
  journal	= {Biometrika},
  volume	= {106},
  number	= {2},
  pages		= {479--486},
  year		= {2019},
  publisher	= {Oxford University Press}
}

@Article{matsubara2022generalised,
  title		= {Generalised {B}ayesian Inference for Discrete Intractable
		  Likelihood},
  author	= {Matsubara, Takuo and Knoblauch, Jeremias and Briol,
		  Fran{\c{c}}ois-Xavier and Oates, Chris and others},
  journal	= {Journal of the American Statistical Association},
  volume	= {119},
  number	= {547},
  pages		= {2345--2355},
  year		= {2024}
}

@article{rizzo2016energy,
  title={Energy distance},
  author={Rizzo, Maria L and Sz{\'e}kely, G{\'a}bor J},
  journal={wiley interdisciplinary reviews: Computational statistics},
  volume={8},
  number={1},
  pages={27--38},
  year={2016},
  publisher={Wiley Online Library}
}

@article{johndrow2020scalable,
  title={Scalable approximate {MCMC} algorithms for the horseshoe prior},
  author={Johndrow, James and Orenstein, Paulo and Bhattacharya, Anirban},
  journal={Journal of Machine Learning Research},
  volume={21},
  number={73},
  pages={1--61},
  year={2020}
}

@inproceedings{gal2016dropout,
  title={Dropout as a {B}ayesian approximation: Representing model uncertainty in deep learning},
  author={Gal, Yarin and Ghahramani, Zoubin},
  booktitle={international conference on machine learning},
  pages={1050--1059},
  year={2016},
  organization={PMLR}
}

@article{mohammed2023comprehensive,
  title={A comprehensive review on ensemble deep learning: Opportunities and challenges},
  author={Mohammed, Ammar and Kora, Rania},
  journal={Journal of King Saud University-Computer and Information Sciences},
  volume={35},
  number={2},
  pages={757--774},
  year={2023},
  publisher={Elsevier}
}

@article{goodfellow2014generative,
   title={Generative Adversarial Networks},
   author={Goodfellow, Ian J. and Pouget-Abadie, Jean and Mirza, Mehdi and Xu, Bing and Warde-Farley, David and Ozair, Sherjil and Courville, Aaron and Bengio, Yoshua},
   journal={Advances in Neural Information Processing Systems 27 (NeurIPS 2014)},
   pages={2672--2680},
   year={2014}
 }

@inproceedings{adam2014method,
   title={Adam: A Method for Stochastic Optimization},
   author={Kingma, Diederik P and Ba, Jimmy},
   booktitle={International Conference on Learning Representations},
   year={2015}
 }

@article{Beaumont_Zhang_Balding_2002,
   title={Approximate {B}ayesian Computation in Population Genetics},
   author={Beaumont, Mark A and Zhang, Wenyang and Balding, David J},
   journal={Genetics},
   volume={162},
   number={4},
   pages={2025--2035},
   year={2002}
 }

@inproceedings{Lueckmann_Goncalves_Bassetto_Öcal_Nonnenmacher_Macke_2017,
   title={Flexible statistical inference for mechanistic models of neural dynamics},
   author={Lueckmann, Jan-Matthis and Goncalves, Pedro J. and Bassetto, Giacomo and Öcal, Kaan and Nonnenmacher, Marcel and Macke, Jakob H.},
   booktitle={Advances in Neural Information Processing Systems 30 (NeurIPS 2017)},
   year={2017}
 }

@inproceedings{Goodfellow_Pouget-Abadie_Mirza_Xu_Warde-Farley_Ozair_Courville_Bengio_2014,
   title={Generative Adversarial Networks},
   author={Goodfellow, Ian J. and Pouget-Abadie, Jean and Mirza, Mehdi and Xu, Bing and Warde-Farley, David and Ozair, Sherjil and Courville, Aaron and Bengio, Yoshua},
   booktitle={Advances in Neural Information Processing Systems 27 (NeurIPS 2014)},
   pages={2672--2680},
   year={2014}
 }

@misc{Mohamed_Lakshminarayanan_2017,
   title={Learning in Implicit Generative Models},
   author={Mohamed, Shakir and Lakshminarayanan, Balaji},
   eprint={1610.03483},
   archivePrefix={arXiv},
   primaryClass={stat.ML},
   year={2017}
 }

@inproceedings{Nowozin_Cseke_Tomioka_2016,
   title={f-{GAN}: Training Generative Neural Samplers using Variational Divergence Minimization},
   author={Nowozin, Sebastian and Cseke, Botond and Tomioka, Ryota},
   booktitle={Advances in Neural Information Processing Systems 29 (NeurIPS 2016)},
   pages={271--279},
   year={2016}
 }

@inproceedings{Wenzel_Roth_Veeling_Świątkowski_Tran_Mandt_Snoek_Salimans_Jenatton_Nowozin_2020,
   title={How Good is the {B}ayes Posterior in Deep Neural Networks Really?},
   author={Wenzel, Florian and Roth, Kevin and Veeling, Bastiaan S. and Świątkowski, Jakub and Tran, Linh and Mandt, Stephan and Snoek, Jasper and Salimans, Tim and Jenatton, Rodolphe and Nowozin, Sebastian},
   booktitle={Proceedings of the 37th International Conference on Machine Learning},
   volume={119},
   pages={10248--10259},
   year={2020},
   organization={PMLR}
 }

@article{makalic2015simple,
  title={A simple sampler for the horseshoe estimator},
  author={Makalic, Enes and Schmidt, Daniel F},
  journal={IEEE Signal Processing Letters},
  volume={23},
  number={1},
  pages={179--182},
  year={2015},
  publisher={IEEE}
}

@article{ghosh2019model,
  title={Model selection in Bayesian neural networks via horseshoe priors},
  author={Ghosh, Soumya and Yao, Jiayu and Doshi-Velez, Finale},
  journal={Journal of Machine Learning Research},
  volume={20},
  number={182},
  pages={1--46},
  year={2019}
}

@article{bhadra2019lasso,
  title={Lasso meets horseshoe},
  author={Bhadra, Anindya and Datta, Jyotishka and Polson, Nicholas G and Willard, Brandon},
  journal={Statistical Science},
  volume={34},
  number={3},
  pages={405--427},
  year={2019},
  publisher={JSTOR}
}

@article{carvalho2010horseshoe,
  title={The horseshoe estimator for sparse signals},
  author={Carvalho, Carlos M and Polson, Nicholas G and Scott, James G},
  journal={Biometrika},
  pages={465--480},
  year={2010},
  publisher={JSTOR}
}

@article{Sinha_Roy_2026,
   title={Prequential posteriors},
   ISSN={2520-8764},
   url={http://dx.doi.org/10.1007/s42081-026-00374-w},
   DOI={10.1007/s42081-026-00374-w},
   journal={Japanese Journal of Statistics and Data Science},
   publisher={Springer Science and Business Media LLC},
   author={Sinha Roy, Shreya and Everitt, Richard G. and Robert, Christian P. and Dutta, Ritabrata},
   year={2026},
   month=Sept }

@article{ho2020denoising,
  title={Denoising diffusion probabilistic models},
  author={Ho, Jonathan and Jain, Ajay and Abbeel, Pieter},
  journal={Advances in Neural Information Processing Systems},
  volume={33},
  pages={6840--6851},
  year={2020}
}

@article{anand2022protein,
  title={Protein sequence and structure generation with equivariant denoising diffusion probabilistic models},
  author={Anand, Namrata and Achim, Tudor},
  journal={arXiv preprint arXiv:2205.15019},
  year={2022}
}

@article{huang2024harnessing,
  title={Harnessing deep learning for population genetic inference},
  author={Huang, Xin and Rymbekova, Aigerim and Dolgova, Olga and Lao, Oscar and Kuhlwilm, Martin},
  journal={Nature Reviews Genetics},
  volume={25},
  number={1},
  pages={61--78},
  year={2024},
  publisher={Nature Publishing Group UK London}
}

@article{yelmen2023overview,
  title={An overview of deep generative models in functional and evolutionary genomics},
  author={Yelmen, Burak and Jay, Flora},
  journal={Annual Review of Biomedical Data Science},
  volume={6},
  number={1},
  pages={173--189},
  year={2023},
  publisher={Annual Reviews}
}

@article{johnstone2004needles,
  title={Needles and straw in haystacks: Empirical Bayes estimates of possibly sparse sequences},
  author={Johnstone, Iain M and Silverman, Bernard W},
  journal={The Annals of Statistics},
  volume={32},
  number={4},
  pages={1594--1649},
  year={2004},
  publisher={Institute of Mathematical Statistics}
}

@article{castillo2012needles,
  title={Needles and straw in a haystack: Posterior concentration for possibly sparse sequences},
  author={Castillo, Isma{\"e}l and Van Der Vaart, Aad},
  journal={The Annals of Statistics},
  pages={2069--2101},
  year={2012},
  publisher={JSTOR}
}

@article{polson2010shrink,
  title={Shrink globally, act locally: Sparse Bayesian regularization and prediction},
  author={Polson, Nicholas G and Scott, James G},
  journal={Bayesian statistics},
  volume={9},
  number={501-538},
  pages={105},
  year={2010},
  publisher={Oxford}
}

@article{hoerl1970ridge,
  title={Ridge regression: Biased estimation for nonorthogonal problems},
  author={Hoerl, Arthur E and Kennard, Robert W},
  journal={Technometrics},
  volume={12},
  number={1},
  pages={55--67},
  year={1970},
  publisher={Taylor \& Francis}
}

\appendix
\section{Splitting scheme for preconditioned adjusted stochastic gradient Langevin dynamics}
\label{ChapHSsec:splitscheme}
To generate more accurate samples, we resort to an
easily implemented 2nd-order symmetric splitting scheme (SSS).
The idea of SSS is to split an infeasible SDE into several sub-SDEs, with each part analytically solvable;
approximate samples are then generated by sequentially evolving parameters via these sub-SDEs. 
Specifically, we split the SDE in Equation~\eqref{ChapHSeq:adSGLD} into the following three sub-SDEs:
\begin{align*}
	&A: \left\{\begin{array}{ll}
	d \boldsymbol{\theta}&= H(\boldsymbol \theta)\boldsymbol{p} d t,\\
    d \boldsymbol{p}&= 0 \\
    d \boldsymbol{\xi} &= \left(\frac{1}{p}\boldsymbol{p}^T\boldsymbol{p} - 1\right)d t
	\end{array}\right.,
	B: \left\{\begin{array}{ll}
	d \boldsymbol{\theta}&= 0 \\
    d \boldsymbol{p}&=-\boldsymbol{\xi}\boldsymbol{p}d t \\
    d \boldsymbol{\xi} &= 0
	\end{array}\right. \nonumber 
\end{align*}

\begin{align*}
    &O: \left\{\begin{array}{ll}
	d \boldsymbol{\theta}&= 0\\
    d \boldsymbol{p}&= -H(\boldsymbol{\theta}) \nabla_{\boldsymbol{\theta}} U(\boldsymbol{\theta})d t+\sqrt{2} H(\boldsymbol{\theta})^{\frac{1}{2}} d W\\
    d \boldsymbol{\xi} &= 0
	\end{array}\right. \nonumber
\end{align*}

We then use the order ABOBA to solve the above sub-SDEs to generate approximate
samples. Specifically, based on the splitting, parameters first evolve
following $A$ for time $\eta/2$, then $B$ for time $\eta/2$, $O$ for time $\eta$, 
$B$ again for time $\eta/2$, and finally $A$ again for time $\eta/2$, resulting in 
the update equations in Algorithm~\ref{ChapHSalg:prec-adjSGLD-santa}. The resulting parameters then serve as approximate samples from the target ditribution.

\section{Sensitivity analysis of prior choices for sparse linear regression}\label{ChapHSappendix:lreg_bvlaues_llhd}

\begin{figure}[htbp!]
    \centering
    \includegraphics[width=\linewidth]{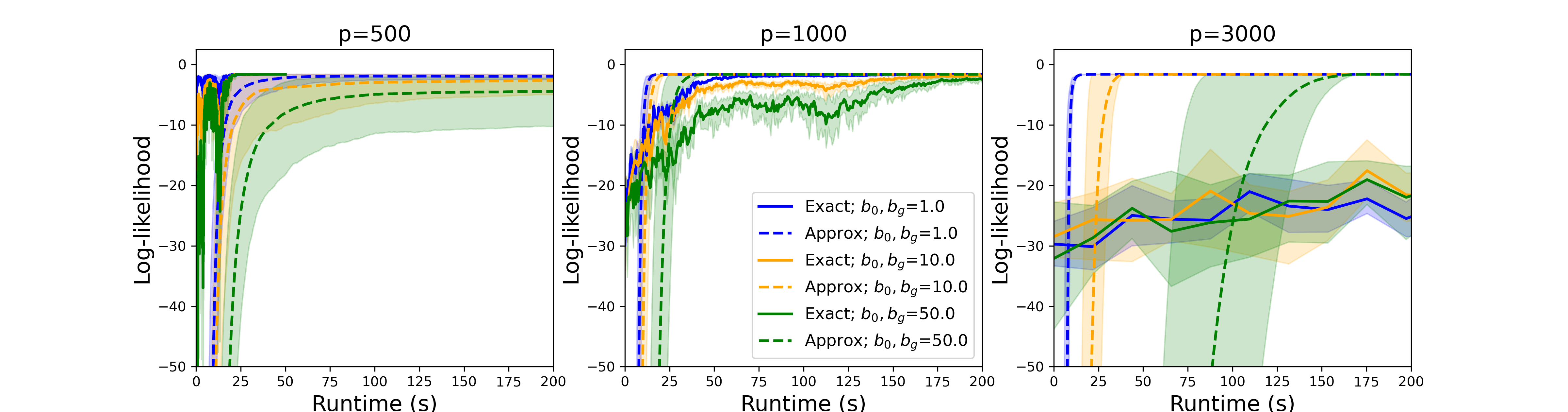}
    \caption{Exact and approximate horseshoe sampler for linear regression with fixed $n=50$. We have considered 3 pairs of different values of the hyperparameters ($b_0, b_g$) to define the horseshoe prior.   }
    \label{ChapHSfig:lreg_differentb_hs}
\end{figure}

We present the impact of the choice of the hyperparameters ($b_0, b_g$) on the exact and the approximate sampler for linear regression in Figure~\ref{ChapHSfig:lreg_differentb_hs}. The choice of hyperparameters has very little impact on the exact sampler; however, the approximate sampler performs the best when $b_0, b_g = 1$.

\section{Sensitivity analysis of prior choices for Hopper}\label{ChapHSappendix:hopper_hs}

For the Hopper experiment we provide the ablation study on $b_0, b_g$ in Table~\ref{ChapHStab:b_blocks_hopper}.

\begin{table}[h!]
\centering
\begin{tabular}{lccc}
\hline
$n$ & NRMSE & $R^2$ & Energy score \\
\hline
\multicolumn{4}{c}{$\mathbf{b_0, b_g = 1/p}$} \\
\hline
50   & 0.543 (0.155) & -27.810 (23.520) & 3.295 (0.155) \\
100  & 0.626 (0.157) & -38.750 (28.170) & 3.407 (0.172) \\
500  & 0.674 (0.189) & -45.410 (30.210) & 3.470 (0.150) \\
1000 & 0.651 (0.219) & -46.080 (31.280) & 3.431 (0.148) \\
\hline
\multicolumn{4}{c}{$\mathbf{b_0, b_g = 0.01}$} \\
\hline
50   & 0.319 (0.136) & -9.580 (11.300)  & 3.046 (0.129) \\
100  & 0.333 (0.135) & -10.990 (12.190) & 3.051 (0.116) \\
500  & 0.281 (0.124) & -8.770 (9.170)   & 2.996 (0.079) \\
1000 & 0.227 (0.145) & -6.600 (9.880)   & 2.982 (0.105) \\
\hline
\multicolumn{4}{c}{$\mathbf{b_0, b_g = 1}$} \\
\hline
50   & 0.212 (0.070) & -3.070 (3.260)   & 3.001 (0.079) \\
100  & 0.215 (0.074) & -3.120 (3.750)   & 2.984 (0.065) \\
500  & 0.152 (0.073) & -1.550 (3.200)   & 2.934 (0.046) \\
1000 & 0.115 (0.055) & -0.470 (1.890)   & 2.914 (0.031) \\
\hline
\multicolumn{4}{c}{$\mathbf{b_0, b_g = 10}$} \\
\hline
50   & 0.204 (0.070) & -3.010 (3.380)   & 2.995 (0.071) \\
100  & 0.202 (0.071) & -2.930 (3.920)   & 2.979 (0.060) \\
500  & 0.135 (0.075) & -1.400 (3.740)   & 2.926 (0.035) \\
1000 & 0.098 (0.061) & -0.320 (2.130)   & 2.908 (0.023) \\
\hline
\multicolumn{4}{c}{$\mathbf{b_0, b_g = 100}$} \\
\hline
50   & 0.205 (0.071) & -3.060 (3.270)   & 2.995 (0.068) \\
100  & 0.204 (0.072) & -3.070 (3.690)   & 2.980 (0.055) \\
500  & 0.154 (0.082) & -1.810 (3.430)   & 2.935 (0.032) \\
1000 & 0.121 (0.079) & -0.740 (2.020)   & 2.916 (0.032) \\
\hline
\end{tabular}
\caption{Predictive performance across different values of the horseshoe hyperparameters $\{b_0,b_g\}$ and observation sizes $n$ for the Hopper experiment. Values are reported as mean (standard deviation) across independent runs.}
\label{ChapHStab:b_blocks_hopper}
\end{table}

We observe similar predictive performance via horseshoe prior when $b_0,b_g \geq 1$. However, the performance slightly deteriorates when the hyperparameters are set to values smaller than 1.

\end{document}